\documentclass[10pt]{spie}  

\usepackage{amsmath,amsfonts,amssymb}
\usepackage{graphicx}
\usepackage[multi-part-units=single]{siunitx}
\usepackage{color}
\usepackage{bm}
\usepackage{url}
\usepackage[export]{adjustbox}
\usepackage{multirow}
\usepackage[colorlinks=true, allcolors=blue]{hyperref}

\title{Melanoma detection with electrical impedance spectroscopy and dermoscopy using joint deep learning models}

\author[a]{Nils Gessert}
\author[a]{Marcel Bengs}
\author[a]{Alexander Schlaefer}
\affil[a]{Institute of Medical Technology, Hamburg University of Technology, Hamburg, Germany}

\authorinfo{Further author information: (Send correspondence to Nils Gessert)\\Nils Gessert: E-mail: nils.gessert@tuhh.de\\ Marcel Bengs: E-mail: marcel.bengs@tuhh.de\\  Alexander Schlaefer: E-mail: schlaefer@tuhh.de}

\begin{document} 
\maketitle
\begin{abstract}
The initial assessment of skin lesions is typically based on dermoscopic images. As this is a difficult and time-consuming task, machine learning methods using dermoscopic images have been proposed to assist human experts. Other approaches have studied electrical impedance spectroscopy (EIS) as a basis for clinical decision support systems. Both methods represent different ways of measuring skin lesion properties as dermoscopy relies on visible light and EIS uses electric currents. Thus, the two methods might carry complementary features for lesion classification. Therefore, we propose joint deep learning models considering both EIS and dermoscopy for melanoma detection. For this purpose, we first study machine learning methods for EIS that incorporate domain knowledge and previously used heuristics into the design process. As a result, we propose a recurrent model with state-max-pooling which automatically learns the relevance of different EIS measurements. Second, we combine this new model with different convolutional neural networks that process dermoscopic images. We study ensembling approaches and also propose a cross-attention module guiding information exchange between the EIS and dermoscopy model. In general, combinations of EIS and dermoscopy clearly outperform models that only use either EIS or dermoscopy. We show that our attention-based, combined model outperforms other models with specificities of $\SI{34.4}{\percent}$ $(\textrm{CI }31.3$-$38.4)$, $\SI{34.7}{\percent}$ $(\textrm{CI }31.0$-$38.8)$ and $\SI{53.7}{\percent}$ $(\textrm{CI }50.1$-$57.6)$ for dermoscopy, EIS and the combined model, respectively, at a clinically relevant sensitivity of $\SI{98}{\percent}$.
\end{abstract}


\section{Introduction}

Melanoma is one of the most lethal types of skin cancer. Late detection of melanoma is associated with very low survival times \cite{tsao2004management}, making tools for early diagnosis vital for patient care. In particular, dermoscopy has significantly improved diagnostic accuracy over examination with the naked eye \cite{carli2004improvement}. Still, melanoma detection remains a challenging task. In general, excision is performed very often which leads to many melanomas being detected (high sensitivity) but also a lot of unnecessary, costly, and invasive procedures (low specificity) \cite{rocha2017analysis}. Therefore, additional tools for melanoma detection have been proposed, such as electrical impedance spectroscopy (EIS) \cite{aaberg2011electrical}. Here, an electrode is placed on the skin lesion and multiple impedance measurements at different frequencies are performed between several bars. Then, a machine learning algorithm uses the complex impedance data as a feature vector to estimate the probability of the lesion being benign or malignant. The method has been shown to be effective in multiple clinical studies  \cite{malvehy2014clinical,mohr2013electrical}. 

Recently, other automatic diagnosis methods have been proposed using dermoscopic images as an input to convolutional neural networks (CNNs) \cite{kawahara2016deep,lopez2017skin}. Also, the ISIC 2018 Skin Lesion Analysis Towards Melanoma Detection Challenge \cite{codella2019skin} resulted in numerous deep learning-based methods for high-performance skin lesion classification \cite{gessert2018skin}. 

So far, machine learning methods for EIS and dermoscopy have been addressed separately, although dermoscopic images are generally acquired before EIS measurements \cite{rocha2017analysis}. EIS and dermoscopy represent very different ways of measuring skin lesion properties. While dermoscopy captures light absorption mostly at the skin surface, EIS provides electrical resistance properties in deeper skin layers. Thus, the two methods might complement each other and fusing them in a single model could improve clinical decision support. Furthermore, generalization of CNNs for dermoscopy across datasets has been shown to be problematic as models tend to overfit specific image datasets \cite{codella2019skin,tschandl2019comparison}. Incorporating EIS could be helpful in this regard, as EIS classifiers have been shown to generalize well across data from different clinical studies \cite{malvehy2014clinical}. 

As a first step, we study deep learning methods for EIS as an alternative to approaches relying on classic models such as SVMs \cite{mohr2013electrical}. We propose a new recurrent model with state-max-pooling which takes domain knowledge on EIS into account. For each lesion, a varying number of EIS measurements is performed, depending on lesion size. In contrast to previous methods, we directly learn the importance of measurements by treating them as arbitrary-length sequences that are fed into recurrent models. 

Second, we explore lesion classification with dermoscopic images from an EIS study using state-of-the-art CNN-based approaches \cite{gessert2018skin,gessert2019skin}. 

Third, we build combined deep learning-based EIS and dermoscopic models. Here, we propose to use a cross-attention mechanism guiding information exchange between the two data sources. We compare this approach to other linear and nonlinear combination methods. For clinical decision support, automatic systems should operate at a very high sensitivity to ensure that critical lesions are not missed \cite{mohr2013electrical}. Therefore, we evaluate all models at a sensitivity of $\SI{98}{\percent}$.

Summarized, our contributions are two-fold. First, we propose a new model for deep learning with EIS by taking domain knowledge into account. Second, we show that combining EIS and dermoscopy with a cross-attention module substantially outperforms other combination approaches and methods using either EIS or dermoscopy. 
\section{Methods}

\subsection{Dataset}

The dataset we use was collected in a previous multicenter clinical trial across Europe and was kindly provided to us by the curators \cite{mohr2013electrical}. The final dataset we use contains $\num{988}$ lesions with one dermoscopic image each and a total of $\num{3131}$ impedance measurements. There are $\num{631}$ benign lesions and $\num{357}$ malignant lesions which include melanoma, other skin cancer or severe dysplastic lesions. The diagnosis was obtained with histopathological evaluation by three pathologists. Due to the small dataset size we use five-fold cross-validation, i.e. we split the $\num{988}$ lesions into five equally sized sets with the same class balance as the full dataset. For validation, we train on three folds, validate on the fourth and leave out the fifth, repeated for each fold. For testing, we train on four folds and evaluate on the previously left out fold, repeated for each fold. In this way, we obtain predictions for all lesions. Our models classify whether a lesion is benign or malignant. For evaluation, we use all benign lesions and all malignant melanomas, i.e. our reported metrics reflect the performance for melanoma detection, the clinically most challenging lesion type. During training, we use all samples for more data variety. 
\begin{figure}[t]
	\centering
	\adjincludegraphics[width=0.9\columnwidth,trim={0 {.17\height} 0 0},clip]{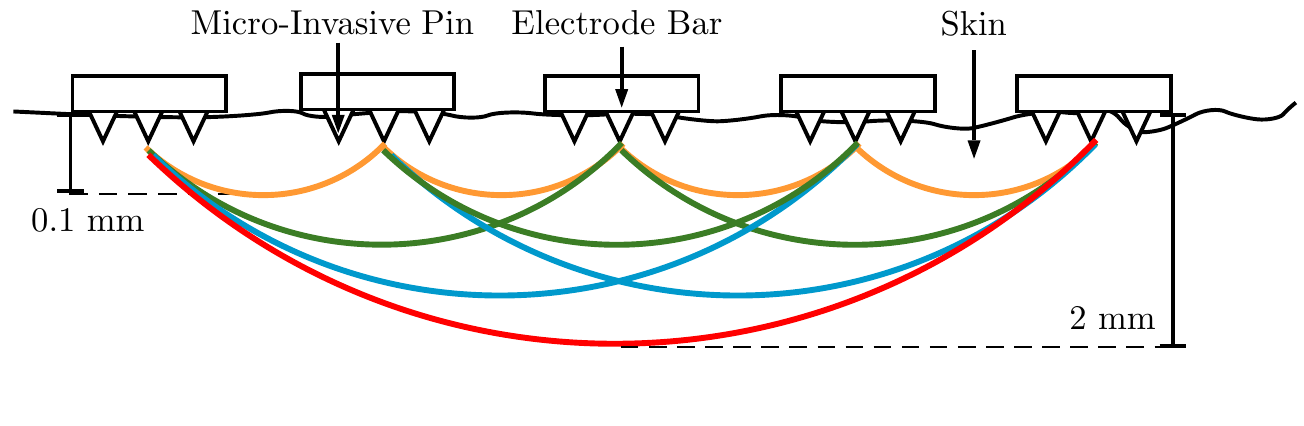}
	\caption{Schematic drawing of the EIS probe, following \cite{mohr2013electrical}. The colors indicate measurements at different depths.}
	\label{fig:EISprobe}
\end{figure}

The EIS device is a Nevisense, manufactured by Scibase AB. The EIS electrode contains five bars. By measuring between different bars, ten different permutations are recorded which correspond to different measurement depths. Measurements between neighboring bars record impedance close to the skin surface while measurements between bars with larger distance correspond to the impedance in deeper skin layers, see Figure~\ref{fig:EISprobe}. For each permutation, frequencies from $\SI{1}{\kilo\hertz}$ to $\SI{2.5}{\mega\hertz}$, distributed on a logarithmic scale, are measured. Thus, the complex impedance values are aggregated in a feature vector of size $700$. 
\begin{figure}[t]
	\centering
	\includegraphics[width=0.45\columnwidth,valign=t]{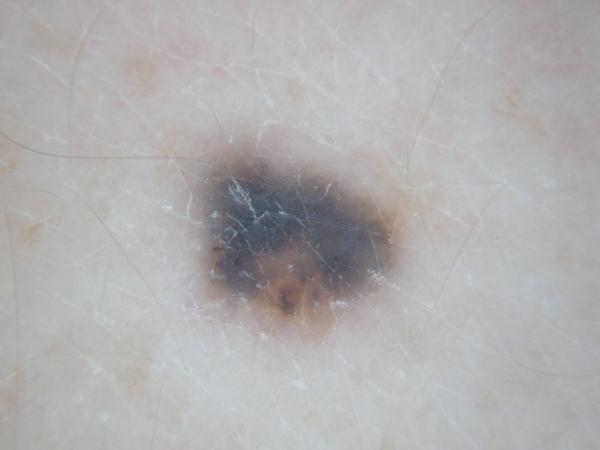}
	\includegraphics[width=0.49\columnwidth,valign=t]{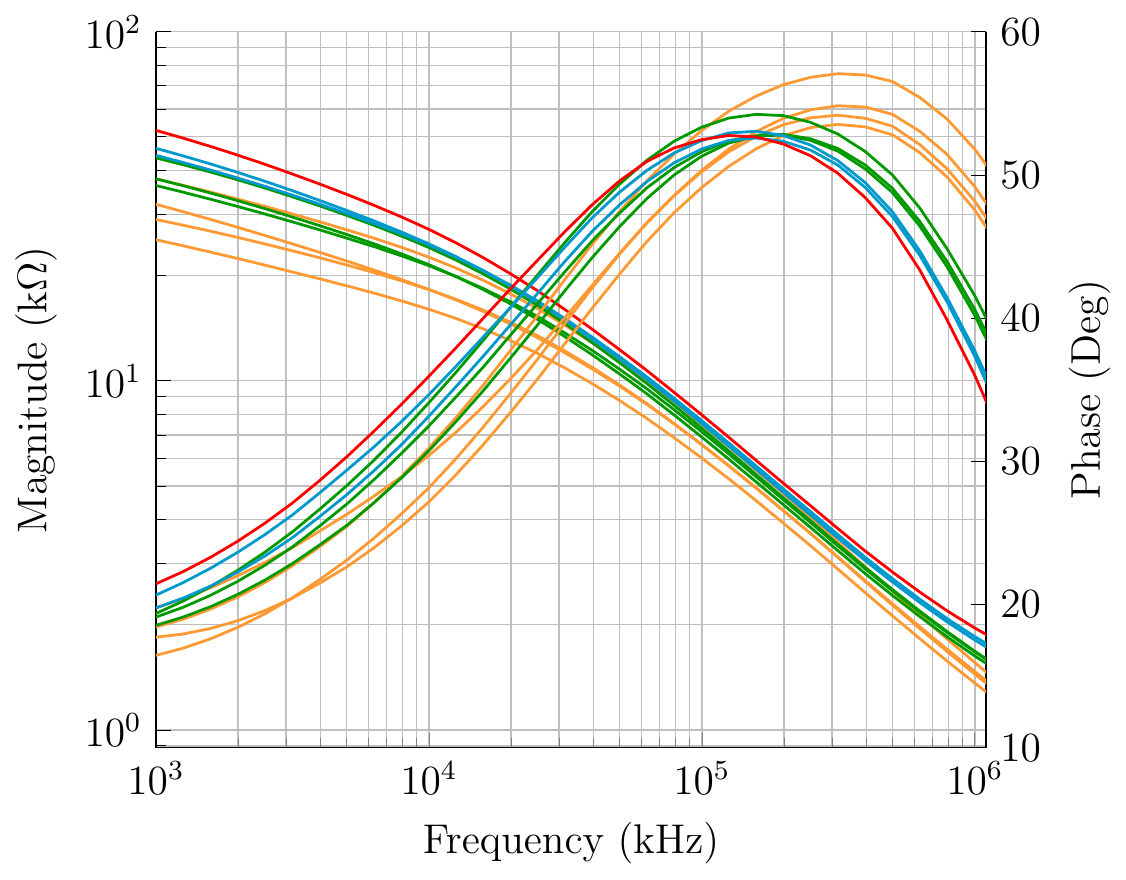}	
	\caption{Example dermoscopy images and impedance data. The complex impedance data is shown as magnitude and phase. Data for a benign lesion is shown. The colors indicate measurements at different depths as indicated in Figure~\ref{fig:EISprobe}.}
	\label{fig:examples}
\end{figure}

The dermoscopic images are of varying size from different devices. First, we manually crop the lesions to cut out zero-signal (black) parts of the image. Then, we apply color constancy to all images using the Shades of Gray method with Minkowski norm $p=6$. Last, we resize the images to $600 \times 450$ pixels, following Tschandl et al. \cite{tschandl2018}. Examples for the dermoscopic images and the EIS data are shown in Figure~\ref{fig:examples}.

\subsection{Models}

\textbf{Baseline Models for EIS Data.} First, we reimplement classical machine learning methods for comparison to previous approaches \cite{mohr2013electrical}. For this purpose we use SVMs and fully-connected neural networks (FC-NNs). The models process one measurement at a time, i.e. for each lesion $i$ and measurement $j$ the model input is a feature vector $x_j^i \in \mathbb{R}^{700}$. We normalize the input data to have zero mean and unit variance. For the SVM, we use a gaussian kernel and a box constraint of $C=1$. For the FC-NN, we use a two-layer network with ReLU activations and batch normalization. 
As most lesions have multiple measurements $j$ for each lesion $i$, previous methods have used heuristics for combination. Similar to previous EIS models, we select the prediction with the highest probability of the lesion being malignant which leads to a high sensitivity of the classifier. Thus, the prediction $p^i$ for lesion $i$ is computed as $p^i = \max \{p^{i}_1, \dots ,p^{i}_{N_i}\}$ where $N_i$ is the number of measurements for lesion $i$. This can be interpreted as max-pooling over the predictions of all measurements for each lesion.

\textbf{Recurrent Aggregation with EIS Data}. Next, we propose a new model to process impedance data. The measurements of each lesion can be interpreted as a sequence of variable length. For this type of data, recurrent models have been very successful \cite{hochreiter1997long}. Therefore, we propose the use of a model with gated recurrent units (GRUs) \cite{cho2014learning}. The GRU receives a sequence $\bm{x} = \{x_1,\dots,x_{N_i}\}$ as its input. The GRU computes a state $h_{j}$ using the input $x_j \in \bm{x}$ and the previous state $h_{j-1}$:
\begin{equation}
\begin{array}{l}
z_j = \sigma (M_z h_{j-1} + L_z x_j) \\
r_j = \sigma (M_r h_{j-1} + L_r x_j) \\
c_j = \tanh (M_c (r_jh_{j-1}) + L_cx_i) \\
h_j = z_j c_j+(1-z_j)h_{j-1}
\end{array}
\end{equation}
where $M$ and $L$ are weight matrices.
A typical approach for GRUs is to use the last state $h_{N_i}$ of the GRU as the output. However, this strategy might not be optimal for this type of data as recurrent models assume ordered sequences while the EIS measurements' order is arbitrary. While the last state does contain gated information from all previous states, the last sequence input $x_{N_{i}}$ has the largest contribution which is not desirable for our problem. Instead, we incorporate the max-pooling heuristic used for the baseline model directly into the GRU. Consider a state matrix $H$ with vectors $h_j \in \{h_1,\dots,h_{N_i}\}$ as its columns, obtained from the input sequence $\bm{x}$. We calculate the final GRU output as $o = \mathit{MAX}(H)$ where $\mathit{MAX}$ is a max pooling operation with a pooling window of size $1\times N_{i}$. We refer to this approach as state-max-pooling and compare to the use of state-average-pooling and using $h_{N_i}$ as the GRU output. After this unit, we feed the vector $o$ into another fully-connected hidden layer with batch normalization and ReLU activation and last, we apply the output layer. Both the FC-NN and GRU models are trained with Adam for $200$ epochs, a batch size of $10$ and a learning rate of $\num{2e-5}$. Since the sequence of EIS measurements is arbitrary, we randomly permute the sequences during training. For evaluation, we average over the result of five random permutations.

\textbf{Models for Dermoscopic Data.} We follow the approach proposed in \cite{gessert2018skin} for CNNs with dermoscopy images. The approach lead to the best performance in the ISIC 2018 Challenge when using only publicly available data and its code is publicly available, allowing for reproducibility in contrast to other approaches. Furthermore, the method performed best for binary lesion classification \cite{tschandl2019comparison} which we address in this work. In detail, we use the state-of-the-art models Densenet121 and SE-Resnext50. Due to small dataset size we consider the typical approach of pretraining on ImageNet in comparison to training from scratch. The last layer is replaced with a layer for binary classification. During training, we use random crops of size $224\times 224$ and data augmentation with random flipping and random contrast and brightness changes. For evaluation, we use multi-crop evaluation with $36$ evenly spread crops across the image which covers the entire image with large overlap. The final prediction is obtained by averaging over the crops' individual predictions.
We train the models with Adam for $100$ epochs, a batch size of $20$ and a learning rate of $\num{2.5e-5}$.
\begin{figure}[t]
	\centering
	\includegraphics[width=1.0\columnwidth]{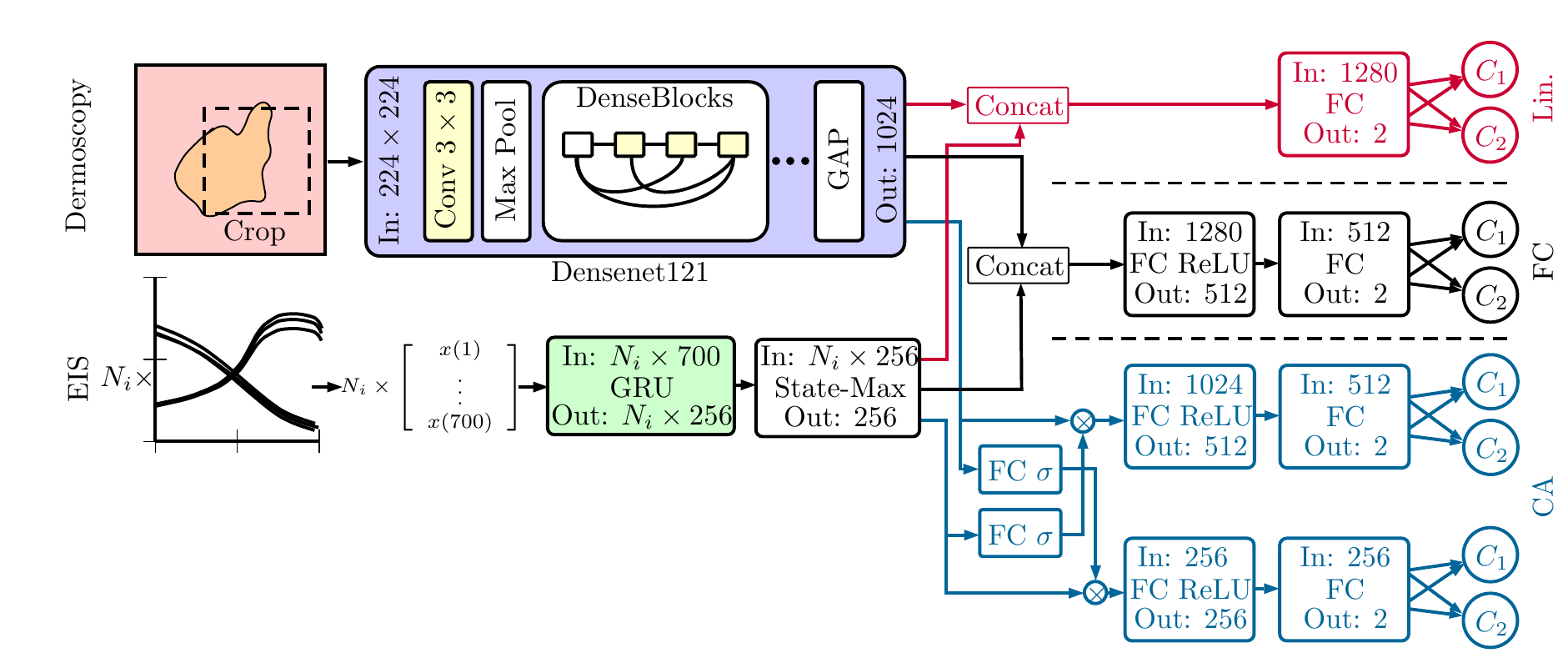}
	\caption{The joint models we propose. State-Max refers to our state-max-pooling strategy. FC denotes a fully-connected layer. GAP denotes global average pooling. Linear combination of features is shown in red (Lin.). Nonlinear combination of features with an FC-layer is shown in black (FC). Our proposed cross-attention mechanism is shown in blue (CA). }
	\label{fig:model}
\end{figure}

\textbf{Combined Models.} As a next step, we combine our models for EIS and dermoscopic data. A straight-forward way of combining models is ensembling. Here, we combine the predictions $p_{\mathit{EIS}}$ and $p_{\mathit{CNN}}$ from the two independent models by selecting the maximum predicted probability $p = \max\{p_{\mathit{EIS}},p_{\mathit{CNN}}\}$ for each lesion. Furthermore, we build joint models where both data sources are fed into the model. In this way, the models can directly learn a combination of features. The joint models are depicted in Figure~\ref{fig:model}. The data sources are first processed independently by the EIS and CNN models introduced above.

Before the output, features from both models are combined in different ways. First, we consider a simple concatenation and linear combination of the features (Dense-GRU Lin.), similar to a previous approach \cite{niu2018multi}. Second, we provide the model with more power by adding an FC layer after concatenation which allows the model to learn a nonlinear combination of the features (Dense-GRU FC). Third, we do not combine the features explicitly but instead let them learn interactions between each other by employing a cross-attention module (Dense-GRU CA). Here, we use an FC layer with sigmoid activation to learn a weighting for the CNN features using the EIS features and vice versa. Each path has an individual output and we obtain predictions by selecting the maximum output value, similar to the ensembling strategy.

The model is trained end-to-end. For the CNN path we use Densenet121 and SE-Resnext50. For the EIS path, we use our novel recurrent approach. During training, a single crop of size $224\times 224$ and a random permutation of the EIS sequence is fed into the model in each iteration. For evaluation we also use the multi-crop evaluation strategy which covers the entire dermoscopy image. We use a random permutation of the EIS sequence for each of the $36$ crops from the dermoscopy images. Again, the final prediction is obtained by averaging the probabilities from all crops.

\section{Results}

\begin{table}[!t]
\setlength{\tabcolsep}{12pt}
\caption{The results of our experiments, given in percent. We show results for using EIS only, dermoscopy (Derm.) only and combined models. The CNNs are Densenet121 (Dense) and SE-Resnext50 (SE-RX). The \SI{95}{\percent} confidence intervals are provided in brackets.}
\label{tab:res}
\centering
\begin{tabular}{l l l l l}
 & & Sensitivity & Specificity & AUC  \\
\hline
\parbox[t]{2mm}{\multirow{5}{*}{\rotatebox[origin=c]{90}{EIS}}} & SVM & $98.0$ $(95.1$-$99.5)$ & $27.9$ $(24.7$-$31.3)$ & $80.8$ $(77.3$-$84.0)$ \\
& FC-NN & $98.0$ $(95.3$-$99.5)$ & $30.0$ $(26.9$-$34.2)$ & $81.9$ $(78.5$-$84.8)$ \\
& GRU last state & $98.0$ $(95.3$-$99.5)$ & $25.0$ $(21.9$-$28.5)$ & $83.9$ $(80.6$-$86.6)$ \\
& GRU state-mean-pool & $98.0$ $(95.3$-$99.5)$ & $25.8$ $(22.6$-$29.1)$ & $83.1$ $(80.1$-$86.0)$ \\
& \textbf{GRU state-max-pool} & $98.0$ $(95.4$-$99.5)$ & $\pmb{34.7}$ $(31.0$-$38.8)$ & $\pmb{84.4}$ $(81.3$-$87.1)$ \\
\hline
\parbox[t]{2mm}{\multirow{4}{*}{\rotatebox[origin=c]{90}{Derm.}}} & Dense & $98.0$ $(95.4$-$99.5)$ & $21.3$ $(18.4$-$24.9)$ & $83.9$ $(80.2$-$86.7)$  \\
& \textbf{Dense Pre} & $98.0$ $(94.8$-$99.5)$ & $\pmb{34.4}$ $(31.3$-$38.4)$ & $88.8$ $(85.6$-$91.0)$ \\
& SE-RX & $98.0$ $(95.3$-$99.5)$ & $17.9$ $(15.1$-$21.1)$ & $83.5$ $(80.1$-$86.8)$ \\
& \textbf{SE-RX Pre} & $98.0$ $(95.1$-$99.5)$ & $34.2$ $(30.3$-$37.9)$ & $\pmb{88.9}$ $(86.0$-$90.9)$ \\
\hline
\parbox[t]{2mm}{\multirow{4}{*}{\rotatebox[origin=c]{90}{Ensemble}}} & SE-RX \& Dense & $98.0$ $(95.2$-$99.5)$ & $31.9$ $(28.6$-$35.8)$ & $89.9$ $(82.2$-$92.1)$ \\
 & GRU \& FC-NN & $98.0$ $(95.1$-$99.5)$ & $32.9$ $(29.4$-$36.5)$ & $82.3$ $(78.9$-$85.2)$ \\
 & \textbf{Dense \& GRU} & $98.0$ $(95.2$-$99.5)$ & $39.1$ $(34.4$-$42.6)$ & $\pmb{90.6}$ $(88.0$-$92.7)$ \\
 & \textbf{SE-RX \& GRU} & $98.0$ $(95.3$-$99.5)$ & $\pmb{41.0}$ $(37.5$-$44.8)$ & $88.9$ $(86.1$-$91.3)$ \\
\hline
\parbox[t]{2mm}{\multirow{6}{*}{\rotatebox[origin=c]{90}{Combined}}} & Dense-GRU Lin. & $98.0$ $(95.2$-$99.5)$ & $47.2$ $(43.6$-$51.5)$ & $90.0$ $(87.3$-$92.0)$ \\
 & SE-RX-GRU Lin. & $98.0$ $(95.3$-$99.5)$ & $42.9$ $(39.0$-$46.6)$ & $90.6$ $(87.9$-$92.6)$ \\
 & Dense-GRU FC & $98.0$ $(95.2$-$99.5)$ & $50.0$ $(47.0$-$53.3)$ & $90.5$ $(88.1$-$92.0)$ \\
 & SE-RX-GRU FC & $98.0$ $(95.1$-$99.5)$ & $52.1$ $(48.3$-$55.8)$ & $90.5$ $(88.1$-$92.4)$ \\
 & Dense-GRU CA & $98.0$ $(95.3$-$99.5)$ & $52.3$ $(48.6$-$56.2)$ & $90.6$ $(88.2$-$92.6)$ \\
 & \textbf{SE-RX-GRU CA} & $98.0$ $(95.1$-$99.5)$ & $\pmb{53.7}$ $(50.1$-$57.6)$ & $\pmb{91.2}$ $(89.0$-$93.0)$ \\

\hline
\end{tabular}
\end{table}

We evaluate at a high, clinically relevant sensitivity of $\approx \SI{98}{\percent}$ by manually choosing a suitable threshold for the predicted probabilities. Above the threshold, a prediction is assigned to the class malignant. For a threshold-independent metric, we consider the area under the curve (AUC). For all metrics we also provide $\SI{95}{\percent}$ confidence intervals (CI) which are obtained by bias corrected and accelerated bootstrapping with $n_{\mathit{CI}} = \num{10000}$ samples. Furthermore, we test for statistical significance in terms of the models' specificity using a permutation test with $n_{\mathit{P}} = \num{10000}$ permutations \cite{efron1994introduction} and a significance level of $\alpha = 0.05$.

The results are shown in Table~\ref{tab:res}. For EIS, our novel GRU-based state-max-pooling model performs best. Without state-max-pooling, the GRU's specificity is lower than the classic models' result. In particular, the performance difference between the GRU with state-max-pooling and all other EIS methods is significant in terms of the specificity. For dermoscopy-based melanoma detection with CNNs, pretraining substantially improves performance. Comparing methods for EIS and dermoscopy, both perform similar. The specificity is not significantly different between the best performing EIS model and the best performing CNN model. When combining CNN and EIS models by ensembling, performance is substantially improved. In contrast, the combination of two dermoscopy or two EIS models by ensembling does not lead to improved performance. The specificity is significantly different between the EIS \& CNN ensembles and the ensembles with the same data source. Explicitly learning from both data sources in a joint model (Combined) improves performance even further with our proposed attention mechanism performing best. The combined models with FC and CA data fusion significantly outperform all ensembles in terms of the specificity.


\section{Discussion}

We address melanoma detection using both EIS and dermoscopy data with deep learning methods. A clinically applicable diagnostic tool is required to have a very high sensitivity as missed out malignant lesions can impact patient survival \cite{tsao2004management}. Therefore, we evaluate our models at $\SI{98.0}{\percent}$ sensitivity which was deemed useful in a previous study \cite{mohr2013electrical} and we assess their specificity at this operating point. Thus, clinically useful models achieve a high specificity given a fixed, high sensitivity. 

For construction of our joint model, we first revisit machine learning methods for EIS data. Our novel GRU-based approach significantly outperforms SVMs and FC-NNs with a specificity of $\SI{34.7}{\percent}$ $(\textrm{CI }31.0$-$38.8)$ compared to $\SI{27.9}{\percent}$ $(\textrm{CI }24.7$-$31.3)$ and $\SI{30.0}{\percent}$ $(\textrm{CI }26.9$-$34.2)$. The GRU-based model takes the nature of EIS data into account as it is able to directly learn which measurements are important, given a sequence of EIS measurements with arbitrary length. The state-max-pooling mechanism integrates previously used heuristics \cite{mohr2013electrical} for high-sensitivity classifier design smoothly into our model.

Second, we perform melanoma detection using only CNNs with the dermoscopic images from the dataset. Comparing EIS and dermoscopy models, the performance in terms of the specificity is not significantly different while the AUC is slightly higher for the CNNs. However, when restricting the models to the same dataset without pretraining, the EIS models perform better with a specificity of $\SI{34.7}{\percent}$ $(\textrm{CI }31.0$-$38.8)$ compared to $\SI{21.3}{\percent}$ $(\textrm{CI }18.4$-$24.9)$. This indicates that EIS models perform better when training from scratch with a small dataset.

Third, we combine both data sources to further improve performance. When taking the simple approach of ensembling the models, the clinically relevant operating point shows a substantially improved specificity over single models. Critically, this kind of performance improvement cannot be observed when combining two models with the same data source, i.e., two CNNs or two EIS-based models. Thus, EIS and dermoscopy appear to complement each other well for melanoma detection. This matches the expectation that EIS and dermoscopy carry complementary features as EIS captures skin properties in deeper layers while dermoscopy mostly provides information at the skin surface.

When building a joint model, the specificity improves even further while the AUC also improves slightly. It is notable that an additional FC layer after feature concatenation improves performance. The more powerful nonlinear transformation most likely allows the model to learn a more meaningful combination of the features. Performance improves further if we do not combine both data sources explicitly, but instead let them learn their interaction with a cross-attention module. The two nonlinear feature transformations FC and CA both significantly improve performance over all ensemble techniques. Overall, we achieve a specificity of $\SI{53.7}{\percent}$ $(\textrm{CI }50.1$-$57.6)$ compared to a specificity of $\SI{34.7}{\percent}$ $(\textrm{CI }31.0$-$38.8)$ for EIS only and $\SI{34.4}{\percent}$ $(\textrm{CI }31.3$-$38.4)$ for dermoscopy only. Thus our combination of EIS and dermoscopy leads to a significant improvement over the current state-of-the-art approach of using dermoscopy and EIS separately.


\section{Conclusion}

We propose to combine electrical impedance spectroscopy (EIS) and dermoscopy in joint deep learning models for improved melanoma detection. For this purpose, we first design a new deep learning model for EIS by using recurrent architectures with state-max-pooling for automatic selection of the most relevant EIS measurement. Second, we fuse this model with convolutional neural networks for dermoscopic image processing and study different ways of combining the two data sources. We find that joint nonlinear feature transformations perform best and we show that our combined approach significantly outperforms models that only use one data source. Our results imply that EIS and dermoscopy carry complementary features that can be effectively exploited by joint deep learning methods. For future work, our approach could be evaluated on larger datasets and its value for clinical decision support could be studied.

\section*{ACKNOWLEDGMENTS} 

This work was partially supported by the TUHH $i^3$ initiative.

\bibliographystyle{spiebib} 
\bibliography{egbib} 

\end{document}